%% file: main.tex
\documentclass[letterpaper]{article} 
\usepackage[preprint]{aaai2027}  
\usepackage[hyphens]{url}  
\usepackage{graphicx} 
\usepackage{natbib}  
\usepackage{caption} 
\usepackage{subcaption}
\usepackage{algorithm}
\usepackage{algorithmic}

\usepackage{newfloat}
\usepackage{listings}
\DeclareCaptionStyle{ruled}{labelfont=normalfont,labelsep=colon,strut=off} 
\floatstyle{ruled}
\newfloat{listing}{tb}{lst}{}
\floatname{listing}{Listing}

\usepackage{booktabs}

\usepackage{subcaption}

\usepackage{booktabs}
\usepackage{amsmath}
\usepackage{amssymb}
\usepackage{array}
\usepackage{multirow}
\usepackage{arydshln} 
\usepackage{algorithm}
\usepackage{algorithmic}
\usepackage{xcolor} 

\usepackage{tcolorbox}
\tcbuselibrary{listings,breakable,skins}

\lstdefinestyle{appendixpython}{
    language=Python,
    basicstyle=\ttfamily\footnotesize,
    keywordstyle=\color{blue!70!black},
    stringstyle=\color{red!65!black},
    commentstyle=\color{green!40!black},
    numbers=none,
    numberstyle=\footnotesize,
    numbersep=8pt,
    showstringspaces=false,
    breaklines=true,
    breakatwhitespace=false,
    columns=fullflexible,
    keepspaces=true,
    frame=none,
    xleftmargin=0pt,
    xrightmargin=0pt
}

\newtcblisting{appendixcode}[1]{%
    enhanced,
    breakable,
    listing only,
    listing engine=listings,
    title={#1},
    width=\linewidth,
    colback=gray!10,
    colframe=black!70,
    colbacktitle=black!75,
    coltitle=white,
    fonttitle=\bfseries,
    boxrule=0.8pt,
    arc=1mm,
    left=4pt,
    right=4pt,
    top=6pt,
    bottom=6pt,
    before skip=4pt,
    after skip=6pt,
    listing options={style=appendixpython}
}

\tcbuselibrary{skins,breakable,listings}
\definecolor{promptText}{HTML}{9B2B2B}
\definecolor{promptHdr}{HTML}{3A3A3A}
\definecolor{promptBg}{HTML}{F0F0F0}
\definecolor{promptHook}{HTML}{9A9A9A}
\lstdefinestyle{graftprompt}{%
  basicstyle=\ttfamily\color{promptText},
  numbers=none,
  breaklines=true, breakatwhitespace=true,
  breakindent=0pt, breakautoindent=false,
  columns=fullflexible, keepspaces=true, showstringspaces=false,
  postbreak=\mbox{\textcolor{promptHook}{$\hookrightarrow$}\space},
  literate={—}{{-{}-}}1 {’}{{'}}1 {‘}{{`}}1 {“}{{"}}1 {”}{{"}}1 {…}{{...}}1,
}
\newtcblisting{promptbox}[1]{%
  breakable, enhanced, arc=3pt, boxrule=0.3pt,
  colback=promptBg, colframe=promptBg,
  colbacktitle=promptHdr, coltitle=white, fonttitle=\bfseries,
  title={#1}, toptitle=3pt, bottomtitle=3pt,
  listing only, listing options={style=graftprompt},
  left=1pt, right=7pt, top=7pt, bottom=7pt,
}

\usepackage{kotex}

\newcommand{\graft}{GRAFT{}}
\DeclareMathOperator*{\argmax}{arg\,max}

\title{Global Optimization and Inference-Time Region Grafting for Agentic Workflows}
\author{
    Donghyeok Koh\textsuperscript{\rm 1}\equalcontrib,
    Gyuwan Kim\textsuperscript{\rm 2}\equalcontrib, 
    Jinyeong Bak\textsuperscript{\rm 3}, 
    Seung-Hoon Na\textsuperscript{\rm 4},
    Tao Yang\textsuperscript{\rm 2}, 
    Haneol Jang\textsuperscript{\rm 1}\corresponding, 
    Cheoneum Park\textsuperscript{\rm 1}\corresponding
}

\affiliations {
    \textsuperscript{\rm 1}HBNU,
    \textsuperscript{\rm 2}UCSB, 
    \textsuperscript{\rm 3}SKKU,
    \textsuperscript{\rm 4}UNIST
}

\begin{document}

\maketitle

\begin{abstract}
Recent advances in agentic workflow optimization automate workflow design through task-specific workflow search or input-conditioned architecture selection.
However, they determine the workflow before execution and cannot adapt failed workflow regions using execution-time label-free quality signals.
Naively enabling such inference-time adaptation through whole-workflow re-optimization would be computationally prohibitive.
To tackle this challenge, we introduce GRAFT, which preserves a globally optimized workflow while locally replacing only selected regions for each input.
Without parameter training, GRAFT evaluates region-level alternatives using label-free execution-quality signals and accepts only replacements that improve local quality while preserving workflow-level consistency, thereby enabling instance-wise adaptation without whole-workflow re-optimization.
GRAFT applies without modification across a range of tasks spanning mathematical reasoning, code generation, and multi-hop and knowledge-intensive question answering.
Under matched optimizer and executor settings, it improves over the strongest prior workflow-optimization method, MaAS, by 3.85 points on average.
Replacing only the executor with a stronger model yields further gains without re-optimizing the global workflow.
This suggests that an optimized workflow is not merely a static optimization artifact, but an adaptable execution policy that can evolve with inference-time feedback and stronger executors.
\end{abstract}


\input{body/01_introduction}

\input{body/02_related_work}

\input{body/03_method}

\input{body/04_experiments}

\input{body/analysis}
\input{body/conclusion}
\bibliography{aaai2027}


\end{document}

%% file: body/01_introduction.tex
\section{Introduction}

Advances in Large Language Model (LLM)~\citep{zhao2026surveylargelanguagemodels, minaee2025largelanguagemodelssurvey} have established \textit{agentic workflows}~\citep{xi2023risepotentiallargelanguage, Wang_2024} that combine reasoning~\cite{cot_neurips22}, retrieval~\cite{rag}, planning~\cite{liu2023llmpempoweringlargelanguage}, memory~\cite{Hatalis_Christou_Myers_Jones_Lambert_Amos-Binks_Dannenhauer_Dannenhauer_2024}, reflection~\cite{reflexion_neurips23}, and self-refinement~\cite{selfrefine} as a standard approach to complex tasks~\citep{autogen, camel, metagpt, agentverse}.
Performance and cost can vary substantially depending on which operators are selected and how they are ordered, even when the same underlying LLM is used~\citep{gptswarm,maas}, making the workflow itself an important target for optimization.
However, designing an optimal workflow for each task relies on human expertise and must be repeated whenever the target task changes. 
Accordingly, research on automatically searching for and optimizing agentic workflows has received considerable attention.
ADAS searches over agent designs represented in code~\citep{adas}, AFlow generates task-specific workflows by searching operator graphs using Monte-Carlo Tree Search (MCTS)~\citep{6145622, aflow}, and MaAS learns an agentic supernet represented as a probability distribution over operators and samples a structure for each query~\citep{maas}.

Previous methods, such as AFlow~\citep{aflow}, commonly search for a single workflow offline, prior to inference, and apply that fixed workflow to every query. 
However, no single fixed structure can be optimal for every input, because inputs differ in difficulty and in the approach they require. 
For example, a simple arithmetic problem needs only a single reasoning step, whereas a multi-step compositional problem requires decomposition, verification, and re-reasoning. 
Because such a workflow is tuned to the dataset average and fixed in advance, it cannot adjust to the difficulty or failure patterns of individual inputs. 
MaAS~\citep{maas} samples a query-specific structure, but from a supernet distribution that is itself learned and then frozen offline. 
DyLAN~\citep{DyLAN} adapts per query through resource allocation and dynamic team formation. 
However, it reconstructs the team for each input instead of building on a task-optimized workflow, trading the stability of a globally validated structure for per-query flexibility. 
Addressing per-input variation therefore calls for re-optimizing the workflow at inference time, once the input is given.

Re-optimizing the structure at inference time involves three challenges:
1) \emph{search space}. The workflow search space spans combinations of operators, parameters, and connections. Existing offline methods invoke an LLM as an optimizer over multiple rounds to explore this space~\citep{adas,aflow}. Repeating the same search for every input is costly.
2) \emph{assessment signal}. Offline search uses ground truth, which is unavailable at inference time.
Without external feedback, having an LLM evaluate and revise its own reasoning can be unreliable and degrade performance.
3) \emph{inter-region interference}.
Failure analyses of multi-agent systems identify misaligned information transfer and error propagation among agents as major failure factors~\citep{mast}.
As workflow stages exchange evidence and intermediate conclusions, locally replacing one component can alter downstream inputs, so improving that component does not guarantee the final outcome.

To tackle these challenges, we introduce \graft, a training-free framework that preserves a globally optimized workflow while adapting only the regions each input requires at inference time, without re-optimizing the whole workflow.
For the search space, \graft{} decomposes a workflow into a sequence of single-entry single-exit (SESE) regions~\citep{sese}, which have been used for the hierarchical decomposition of control-flow graphs in program analysis.
It retains the global workflow searched offline for each task and, at inference time, explores and grafts within a small local candidate space restricted to the role of each region.
As SESE boundaries define the input and output interfaces of a region, internal replacements preserve the workflow connectivity, while semantic effects are controlled through stale propagation.
For the assessment signal, candidates are evaluated without ground truth using a local quality proxy that combines label-free signals, including self-consistency~\citep{selfconsistency} across multiple reasoning chains, groundedness measuring how well the provided evidence supports the answer, and code test pass rates.
To address interference across regions, the coupling guard accepts only replacements that improve local quality and preserve boundary support in evidence-grounded tasks, preventing local grafts from harming global quality.
Winning configurations are stored in memory by input-type signature and reused without search for subsequent inputs, spreading the cost of local search across inputs.
Our contributions are summarized as follows:
\begin{itemize}
\item \textbf{Inference-time adaptation paradigm.} We propose a training-free framework that uses a task-specific globally searched workflow as a complementary starting point and re-optimizes its region-level structure for each input at inference time without gradient updates. It treats a workflow not as a static artifact fixed offline, but as an execution policy that adapts at inference time.
\item \textbf{Label-free local optimization mechanism.}
We design label-free proxies (self-consistency, groundedness, and verifier signals) to evaluate region candidates at inference time without ground truth, a coupling guard that accepts replacement only when boundary consistency is not degraded by $\epsilon$ or more, and a retrieve-or-optimize memory that amortizes the discovery of winning configurations.
\item \textbf{State-of-the-Art performance and executor-transfer} 
When using the same optimizer and executor, \graft{} achieves an average score of 87.44 across five benchmarks, outperforming MaAS by 3.85. 
We separate the contributions of the global workflow source and local grafting through ablations and analyze when label-free proxy fidelity determines performance.
Replacing only the executor with a stronger model at inference time leads to substantial gains, showing that the optimized workflow adapts to stronger executors.
\end{itemize}

%% file: body/02_related_work.tex
\section{Related Work}

\paragraph{Automatic Agentic Workflow Generation.}
Research has explored automatically optimizing the structures and components of LLM-based agents to improve performance.
\citet{adas} employs a meta-agent to iteratively generate and evaluate new agent designs.
\citet{aflow} explores workflows composed of predefined operators using MCTS, and \citet{A2Flow} explores workflows using self-adaptive abstraction operators.
\citet{AgentSquare} and \citet{AgentSwift} search for agent architectures in modular and hierarchical search spaces, respectively, and \citet{FusionFlow} combines workflows discovered through different search processes to generate new workflows.
Prior studies apply a single task-optimized global workflow uniformly to all queries, enabling consistent use of a workflow validated for the task but making it difficult to reflect query-specific difficulty and required reasoning depth. 
\graft{} uses the global workflows as a seed workflow and configures the local workflow differently for each query at inference time.

\paragraph{Supernets and Dynamic Workflow.}
Research has explored query-specific workflows to overcome the limitations of applying a single global workflow to all queries.
\citet{maas} samples a query-specific architecture from a probabilistic agentic supernet. 
\citet{EvoFlow, HFlow} retrieve and evolve workflows using workflow populations and experience memory.
\citet{ScoreFlow} optimizes a workflow generator that constructs a workflow for each task, while \citet{DyLAN} optimizes agent teams using agent importance scores and dynamically reconfigures them during task solving.
\citet{DyFlow} dynamically constructs a workflow based on problem decomposition and intermediate feedback.
\citet{Flow} improves a workflow by adjusting subtask allocations and agent roles during execution.
Existing methods for dynamically constructing query-specific workflow use distributions, generators, or populations learned offline, whereas \graft{} uses a winning configuration library without training and performs inference-time local search.

\paragraph{Prompting and Self-Improvement Operators.}
Methods such as Chain-of-Thought~\cite{cot_neurips22}, Self-Consistency~\cite{selfconsistency}, multi-agent debate~\cite{debate_icml24}, Self-Refine~\cite{selfrefine}, Reflexion~\cite{reflexion_neurips23} have been proposed to improve LLM performance. Retrieval-augmented generation~\cite{rag} and multi-agent frameworks~\cite{autogen} combine these methods.
We use these methods as workflow operators and select their region assignments at both global and local levels.

%% file: body/03_method.tex
\section{Method: \graft}
\input{rsc/architecture}

This section describes \graft. 
We formulate the problem and define the offline global workflow construction, inference-time local region grafting, and the proposer, coupling guard, memory, and inter-region state propagation that coordinate these stages.
Figure~\ref{fig:architecture} illustrates \graft.

\subsection{Preliminaries}

An agentic workflow $W$ is a typed directed acyclic graph (DAG) of operators that generates an output $W(x)$ by sequentially applying the operators to an input $x$.
The operator library $\mathcal{O}$ consists of operators associated with roles such as plan, evidence, reason, verify, refine, and format.
The goal of workflow optimization is a token-regularized objective that balances quality $G$ and token cost $c_{\mathrm{tok}}$, with token cost incorporated into the objective via $\lambda_{\mathrm{tok}}$ rather than imposed as a hard constraint.
\begin{equation}
W_t^{*} = \argmax_{W \in \mathcal{W}_t}\; \big[\, G(W) - \lambda_{\mathrm{tok}}\, c_{\mathrm{tok}}(W) \,\big].
\label{eq:global-obj}
\end{equation}
$W_t^{*}$ denotes the conceptual task-level optimum over the task-specific workflow search space $\mathcal{W}_t$.
The global workflow search space grows exponentially with combinations of operator, parameters, and connections, making re-optimization at inference time costly.
The task-level optimum $W_t^{*}$ provides a prior across the input distribution but does not always coincide with the instance-level optimum $W_x^{*}=\argmax_{W \in \mathcal{W}_t} G(W; x)$ for a given input $x$
The $W_{c^{*}}$ used by \graft{} is an approximation derived from role-scoped candidates. 
It preserves $W_{c^{*}}$ as a global structural prior while locally adapting only the regions that exhibit a mismatch for each input.
\graft{} organizes the fixed workflow $W_{c^{*}}$ shown in Figure~\ref{fig:architecture} as an ordered sequence $R=(g_1,\dots,g_n)$ of single-entry single-exit (SESE) regions~\citep{sese}.
The Router enables multi-pass re-execution of predefined regions without altering the DAG, thereby preserving the topological order.

Each role position corresponds to a graftable region. Assigning an identity operator to an optional role deactivates the corresponding region in the global workflow, as indicated by the dashed boxes in Figure~\ref{fig:architecture}. 
A SESE region has a single-entry boundary and a single-exit boundary, so replacing its internal operator configuration preserves the workflow connection structure and input-output interface.
However, a modified region output may affect the semantic execution of downstream regions, so \graft{} manages this dependency through the coupling guard and stale propagation.
Grafting is structurally local but semantically coupled.
Each region $g_i$ has a local candidate space \textit{region library}, consisting only of operators associated with its role.
The workflow execution state $S$ stores the input $x$ and the outputs generated by the operators.
Each region receives the required state fields as input and updates its predefined output fields.

\subsection{Offline Phase: Per-Task Global Workflow Search}

The global phase is performed once offline to search for a task-specific global workflow $W_{c^{*}}$.
Operators $c=(c_1,\dots,c_n)$ are selected from the region library.
A workflow configuration $c=(c_1,\dots,c_n)$ is constructed by selecting, for each region, an operator $c_i$ from its role-specific region library $C_i$.
Because the region library of each optional role includes an identity operator, the assignment itself determines which regions are activated for a given task.
While the role-specific candidate sets are predefined for each task, both the activation of each role position and its operator configuration are jointly determined through role-scoped search.
We exclude evidence or retrieval roles from math and code tasks that can be solved without external information and include them only for QA tasks, as illustrated by the task-specific active regions on the left side of Figure~\ref{fig:architecture}.
Enforcing the same set of roles across all tasks may cause irrelevant roles to be spuriously selected on a small validation set, thereby degrading performance. We therefore tailor the scope of the candidate sets to each task.
The assignment is selected to maximize an objective that jointly accounts for task performance and execution cost on the validation set $D$.
\begin{equation}
c^{*} = \argmax_{c}\; \frac{1}{|D|} \sum_{(x,y)\in D}
\big[\, s(W_c(x), y) - \lambda\, \tau(W_c(x)) \,\big]
\label{eq:rootstock}
\end{equation}
where $s(\cdot,\cdot)$ denotes the matching score between the reference answer and the generated response, and $\tau(\cdot)$ denotes the number of tokens consumed during execution. 
When multiple configurations achieve the same objective value, we select the one with fewer active regions, thereby removing redundant regions.
The search space is restricted to the Cartesian product of the role-specific candidate sets, allowing the optimal assignment to be identified through exhaustive enumeration or coordinate ascent.
The resulting task-specific workflow $W_{c^{*}}$ serves as the initial workflow for region-level grafting at inference time. 
The proposed method is training-free, without gradient-based updates to model parameters or a separate learned adaptation module.

\subsection{Online Phase: Inference-Time Region Grafting}

Given an input $x$ at inference time, \graft{} constructs an incumbent workflow by locally adjusting only the operators in selected regions while preserving the region order and connectivity of $W_{c^{*}}$
Region$g$ is associated with a local edit space $\mathcal{E}(g)$ consisting of replacement operators with the same role and candidate values for parameters such as the number of self-consistency samples and temperature, whose ranges are expanded only when the incumbent quality falls below the target.
We denote the operator configuration of region $g$ by $\phi$, where $\phi^{\mathrm{inc}}$ is the incumbent configuration and $\phi^{*}$ is the selected configuration.
$W[g{\leftarrow}\phi^{*}]$ denotes the workflow obtained by replacing the configuration of region $g$ with $\phi^{*}$.
To control the search cost, \graft expands the edit space in stages.
It initially evaluates only low-cost candidates close to the incumbent and activates the next stage only when the best candidate fails to reach the target 
$\mathcal{E}_0(g) \subseteq \mathcal{E}_1(g) \subseteq \cdots \subseteq \mathcal{E}_J(g)$.
Each candidate configuration $\phi$ is executed and evaluated using label-free signals. Among the admissible candidates that pass the coupling guard, \graft selects the configuration that maximizes the following objective. 
If no candidate is admissible, the incumbent configuration is retained.
\begin{equation}
\begin{aligned}
\phi^{*} = \argmax_{\phi \in \mathcal{E}(g)}\;& \big[\, Q(\phi; x) - \lambda_{\mathrm{loc}}\, \tau(\phi) \,\big] \\
\text{s.t.}\;\;& \mathrm{accept}(\phi),\;\; \ell(\phi) \le \ell_{\mathrm{budget}}(g).
\end{aligned}
\label{eq:graft-obj}
\end{equation}
$Q_{t,r}$ is a local quality score computed from label-free proxy signals.
For task $t$ and role $r$, it is defined as the weighted average over the set of available signals $\mathcal{K}_{t,r}$.
\begin{equation}
Q_{t,r}(\phi; x) = \frac{\sum_{k \in \mathcal{K}_{t,r}} \alpha_{k}\, q_{k}(\phi; x)}{\sum_{k \in \mathcal{K}_{t,r}} \alpha_{k}}.
\label{eq:local-quality}
\end{equation}
Each signal $q_k \in [0,1]$ serves as a label-free proxy, such as answer agreement across multiple reasoning chains used in self-consistency~\citep{selfconsistency}, groundedness measuring the extent to which an answer is supported by the provided evidence, and verifier signals including the code test pass rate or an LLM-based judgment.
The signal set $\mathcal{K}_{t,r}$ varies across tasks and roles, whereas the weights $\alpha_k$ are fixed constants shared across all tasks.
Since $Q_{t,r}$ is computed as a weighted average over the available signals, it remains within $[0,1]$ regardless of the number of signals.
Groundedness measures the proportion of content words in the set $\mathcal{A}$, extracted from the candidate output, that are covered by the evidence set $E$.
\begin{equation}
q_{\mathrm{gr}} = \frac{1}{|\mathcal{A}|} \sum_{a \in \mathcal{A}} \mathrm{support}(a, E)
\label{eq:groundedness}
\end{equation}
where $\mathrm{support}(a,E) \in \{0,1\}$ indicates whether content word $a$ appears lexically in the evidence set $E$.
This content-word coverage score is used as the groundedness term in Equation~\ref{eq:local-quality}.
In \graft, groundedness is therefore defined as lexical evidence coverage.

Equation~\ref{eq:graft-obj} searches over a compact local search space $\mathcal{E}(g)$ associated with a single region, enabling local optimization to be performed online for each input at inference time.
The term $\ell(\phi)$ denotes the estimated execution latency of candidate configuration $\phi$, while $\ell_{\mathrm{budget}}(g)$ denotes the latency budget assigned to region. Candidates predicted to exceed this budget are filtered out prior to execution, thereby avoiding unnecessary inference cost.

Operator replacement is governed by the coupling guard.
\begin{equation}
\begin{aligned}
\mathrm{accept}(\phi) \iff\; & Q(\phi) > Q(\phi^{\mathrm{inc}}) \;\wedge \\
& \mathrm{Coh}(W[g{\leftarrow}\phi], S) \ge \mathrm{Coh}(W, S) - \epsilon
\end{aligned}
\label{eq:guard}
\end{equation}
$\mathrm{Coh}$ is a boundary-support score that measures the proportion of the boundary trace referenced by the answer that is supported by the evidence set $E$ received from upstream regions.
For tasks without external evidence, $\mathrm{Coh}{=}1$, so the boundary-support condition is automatically satisfied and the guard operates solely based on local quality improvement. 
For tasks requiring evidence, the guard additionally evaluates the boundary-support condition.
Replacements that degrade boundary support by more than $\epsilon$ are rejected even if they improve local quality, thus limiting performance degradation caused by interference across regions.
Since $Q(\phi) > Q(\phi^{\mathrm{inc}})$ is a strict inequality, candidates with equal or lower scores are not accepted.

\subsection{Inference Algorithm}

Searching every region for every input is inefficient. The proposer therefore uses a UCB-based priority~\citep{ucb} to determine which regions should be optimized first.
\begin{equation}
\begin{aligned}
\mathrm{score}(g) &= \hat{v}(g) + U(g) + \rho\,\mathrm{Freq}(g)\,\mathrm{Mod}(g), \\
\hat{v}(g) &= w_q(1{-}Q_g) + w_f\,\mathrm{Fail}(g) + w_m\,\mathrm{Mod}(g)
\end{aligned}
\label{eq:proposer}
\end{equation}
$\hat{v}(g)$ measures the adaptation priority of region $g$ by combining its quality deficit $1-Q_g$, failure frequency $\mathrm{Fail}(g)$, and role modifiability $\mathrm{Mod}(g)$.
$U(g)$ is a UCB exploration bonus that favors less frequently optimized regions.
The third term serves as a proactive prior that prioritizes regions that are both frequently modified and amenable to local adaptation.

\input{rsc/grafting-algo}

Algorithm~\ref{alg:graft} details the per-input inference procedure (Fig.~\ref{fig:architecture}b).
For each frontier region processed in upstream-first order, \graft{} first attempts to reuse a configuration retrieved from memory using the signature $\sigma$.
Local search over $\mathcal{E}(g)$ (Eq.~\ref{eq:graft-obj}) is performed only when no matching configuration is available or the retrieved configuration fails to satisfy the target quality $q^{\ast}$.
Accepted configurations update both the workflow and memory, while downstream regions are marked stale for reevaluation in the next pass.
The procedure terminates once no configuration changes, no stale region remains, and every critical region satisfies the quality target, after which the final answer is read from the execution state $S$.
The signature $\sigma$ serves as the lookup key for configuration memory, enabling \graft{} to reuse previously successful configurations and avoid repeated local search for similar inputs.

The maximum number of passes $P$ determines the propagation depth of an upstream replacement through downstream regions.
A single pass is sufficient for tasks that converge after one round of adaptation, whereas multi-hop tasks that aggregate evidence across multiple regions may benefit from additional passes.
Rather than fixing $P$ as a global constant, we select it separately for each task on the validation set using the same offline search protocol as $W_{c^{*}}$, with $P\in\{1,2,3\}$.

\paragraph{Complexity.}
Instead of re-optimizing the entire role-scoped workflow search space $|\mathcal{C}|=\prod_{i=1}^{n}|C_i|$ for every input, \graft{} 
evaluates only the $K$ regions included in the frontier at each pass and at most $M=\max_i|\mathcal{E}(g_i)|$ candidate configurations per region.
Excluding downstream re-execution, the number of candidate evaluations per input is therefore bounded by $O(P\,K\,M)$, where $K\le n$ and $M\ll|\mathcal{C}|$.
A memory hit bypasses local search for the corresponding region, so the effective number of candidate evaluations decreases as successful configurations accumulate across inputs.

%% file: rsc/architecture.tex
\begin{figure*}[!t]
    \centering
    \includegraphics[width=0.91\linewidth]{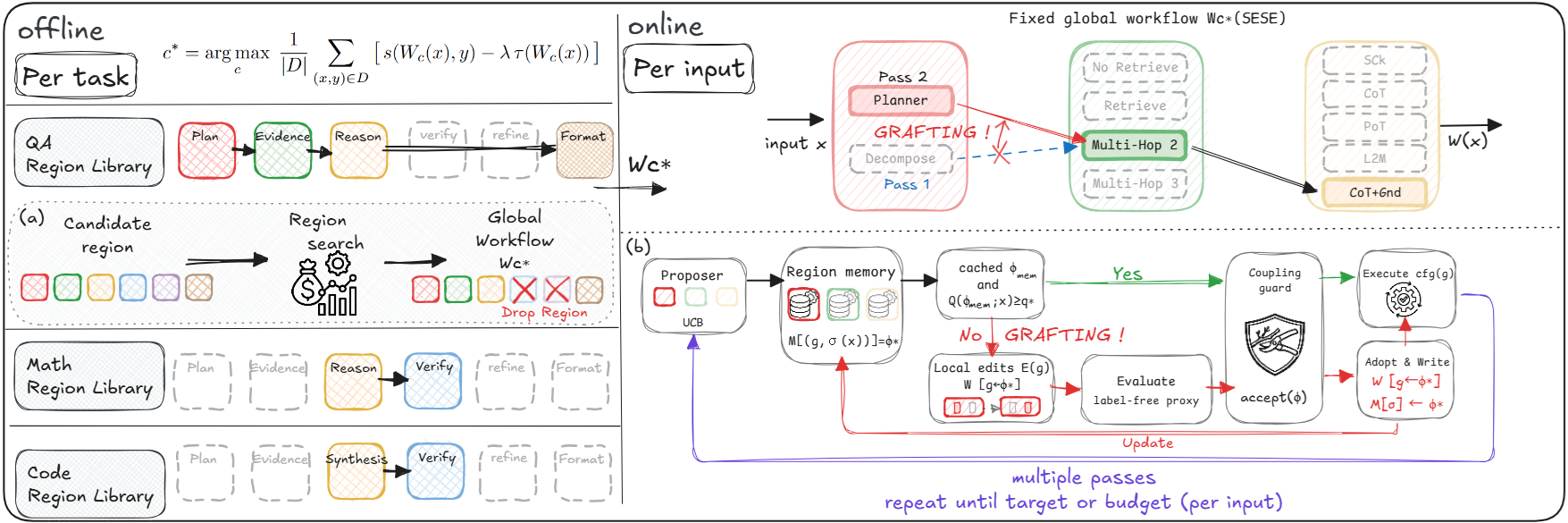}
    \caption{Overview of GRAFT.
    Left: the offline-searched global workflow $W_{c^{*}}$ (dashed boxes are inactive regions). Right: inference-time local grafting for each input (dashed boxes are inactive operators)
    }
    \label{fig:architecture}
\end{figure*}

%% file: rsc/grafting-algo.tex
\begin{algorithm}[tb]
\small
\caption{\graft{} inference (per input $x$)}
\label{alg:graft}
\textbf{Input}: input $x$; task-searched global workflow $W_{c^{*}}$; configuration memory
$\mathcal{M}$; proposer statistics $\Theta$\\
\textbf{Param}: max passes $P$; guard slack $\epsilon$; quality target $q^{\ast}$;
cost budget $B$\\
\textbf{Output}: answer $\hat{a}$
\begin{algorithmic}[1]
\STATE $W \leftarrow W_{c^{*}}$; \; $S \leftarrow \varnothing$; \; $\forall g\!:\; \mathrm{cfg}(g) \leftarrow \mathrm{default}_{W_{c^{*}}}(g)$ \COMMENT{incumbents}
\FOR{$\mathrm{pass} = 1$ \TO $P$}
  \STATE $F \leftarrow \textsc{Propose}(W, S, \Theta)$; \; $\mathit{changed} \leftarrow \textbf{false}$ \COMMENT{eq~\ref{eq:proposer}}
  \FORALL{region $g \in F$ (upstream-first)}
    \IF{cost spent $> B$} \STATE \textbf{break} \ENDIF
    \STATE $\sigma \leftarrow \textsc{Signature}(g, S, x)$ \COMMENT{memory key}
    \STATE $\phi^{*} \leftarrow \mathcal{M}[\sigma]$ \COMMENT{value: past winner}
    \IF{$\phi^{*} = \varnothing$ \textbf{ or } $Q(\phi^{*}; x) < q^{\ast}$}
      \STATE $\phi^{*} \leftarrow \displaystyle\argmax_{\phi \in \mathcal{E}(g)} \big[Q(\phi; x) - \lambda_{\mathrm{loc}}\tau(\phi)\big]$ \COMMENT{eq~\ref{eq:graft-obj}}
    \ENDIF
    \IF{$\phi^{*} \neq \mathrm{cfg}(g)$ \textbf{ and } $\textsc{Accept}(\phi^{*})$ \COMMENT{eq~\ref{eq:guard}}}
      \STATE $\mathrm{cfg}(g) \leftarrow \phi^{*}$; \; $W \leftarrow W[g \leftarrow \phi^{*}]$; \; $\mathcal{M}[\sigma] \leftarrow \phi^{*}$ \COMMENT{adopt}
      \STATE mark downstream$(g)$ stale; \; $\mathit{changed} \leftarrow \textbf{true}$ \COMMENT{cascade}
    \ENDIF
    \STATE $S \leftarrow \textsc{Execute}(\mathrm{cfg}(g), x, S)$ \COMMENT{always run \& update $S$}
  \ENDFOR
  \STATE $\Theta \leftarrow \textsc{Attribute}(S)$ \COMMENT{update $\hat{v}(g)$}
  \IF{$\neg\,\mathit{changed}$ \textbf{ and } no stale region \textbf{ and } $\min_{g \in \mathrm{core}} Q(g) \ge q^{\ast}$}
    \STATE \textbf{break} \COMMENT{early stop}
  \ENDIF
\ENDFOR
\STATE \textbf{return} $\hat{a} \leftarrow \textsc{Readout}(S)$
\end{algorithmic}
\end{algorithm}

%% file: body/04_experiments.tex
\section{Experiments}
Our experiments are designed to answer the following research questions: 
\begin{itemize}
\item \emph{RQ1.} Does inference-time region grafting improve the workflows optimized offline (e.g., AFlow and MaAS)?
\item \emph{RQ2.} Are the gains attributable to local grafting rather than a particular global workflow?
\item \emph{RQ3.} Under what conditions do label-free proxy signals enable or fail to support local search?
\item \emph{RQ4.} Does configuration memory make inference-time adaptation cost-effective?
\end{itemize}


\subsection{Experimental Setup}

\paragraph{Datasets and Metrics.}
To facilitate direct comparison with prior work, we follow the evaluation protocol of MaAS and BayesFlow~\cite{yuan-etal-2026-bayesflow}. 
We evaluate \graft{} on four categories of benchmarks: mathematical reasoning (GSM8K~\citep{gsm8k}, MATH~\citep{math_ds}, and MultiArith~\citep{multiarith}), code generation (HumanEval~\citep{humaneval} and MBPP~\citep{mbpp}), multi-hop question answering (HotpotQA~\citep{hotpotqa} and DROP~\citep{drop}), and knowledge-intensive multiple-choice question answering (MMLU-Pro~\citep{mmlupro}, and GPQA~\citep{gpqa}).
Performance is evaluated using solve rate, pass@1, F1 score, and accuracy, respectively.

\paragraph{Baselines.}
The comparison baselines depend on the evaluation protocol. 
Under the MaAS setup, we compare against single-agent prompting methods (CoT, SC (CoT×5)), automated workflow optimization methods (ADAS, AgentSquare, and AFlow), and MaAS. 
Under the BayesFlow setup, we compare against the baselines reported in the original paper (IO, CoT, CoT-SC, ADAS, MaAS, AFlow, and BayesFlow). 
Baseline results are taken from the corresponding papers, while \graft{} is evaluated on the same public benchmark splits. 


\paragraph{Implementation Details.}
The global workflow $W_{c^{*}}$ is selected once on the validation set using Eq.~(\ref{eq:rootstock}), and only local grafting (Eqs.~(\ref{eq:graft-obj}) and (\ref{eq:guard})) is performed during inference.
The maximum number of passes $P$ and the use of code repair patterns are selected per task on the validation set; all other hyperparameters are shared across tasks.
We set $\epsilon=0.02$ and $\lambda_{\mathrm{tok}}=\lambda_{\mathrm{loc}}=10^{-5}$, with per-instance token and latency budgets.
Prompts follow the benchmark-specified answer format and are identical across all runs.
We average the results over three runs for $W_{c^{*}}$.

\subsection{Main Results}
\input{tabs/tab_main}

We evaluate \graft{} under the MaAS and BayesFlow setups (RQ1).
The MaAS setup uses \texttt{gpt-4o-mini} as executor, and the BayesFlow setup uses Claude Sonnet.
Table~\ref{tab:main} shows that \graft{} achieves state-of-the-art (SOTA) average performance, scoring 87.44 and outperforming all baselines, including MaAS and AFlow.
The average score exceeds MaAS by 3.85, showing that adding region-level local grafting to a global workflow can outperform training-based methods.
Table~\ref{tab:bayesflow} shows that \graft{} achieves SOTA across all benchmarks, with an average of 84.1. 
This shows that inference-time local grafting is also effective for QA and knowledge-intensive tasks.

\subsection{Component Ablation}
\input{tabs/tab_ablcomp}

Table~\ref{tab:abl-comp} compares component ablations that remove inference-time components from the same $W_{c^{*}}$ and different $P$ strategies (RQ2).
Removing local grafting, which is equivalent to executing $W_{c^{*}}$ unchanged, causes the largest average drop of 4.10, confirming that a substantial portion of the performance gain comes from inference-time local grafting.
Replacing the label-free proxy with random scores causes performance to drop by 3.48, nearly matching the setting without grafting and showing that grafting gains come from the proxy signal.
Removing the coupling guard reduces average performance by 0.70, with varying effects across benchmarks.
Since none of the benchmarks in Table~\ref{tab:abl-comp} uses evidence, the boundary-support term is set to $\mathrm{Coh}{=}1$, so the results reflect the effect of accepting only candidates with improved local quality.
In $P$-strategy comparison, per-task $P^{\ast}$ achieves score of 87.44, outperforming both $P{=}3$ and $P{=}1$ and suggesting that selecting $P$ for each task on the validation set is beneficial.

%% file: tabs/tab_main.tex
\begin{table}[t]
\centering

\begin{subtable}{\columnwidth}
\centering
\setlength{\tabcolsep}{5pt}\small
\resizebox{\columnwidth}{!}{%
\begin{tabular}{lcccccc}
\toprule
Method & GSM8K & MATH & MultiArith & HumanEval & MBPP & Avg. \\
\midrule
Vanilla        & 87.45 & 46.29 & 96.85 & 87.08 & 71.83 & 77.50 \\
CoT            & 87.10 & 46.40 & 96.31 & 88.13 & 71.83 & 77.95 \\
SC (CoT$\times$5) & 87.57 & 47.91 & 96.58 & 88.60 & 73.60 & 78.85 \\
ADAS           & 86.12 & 43.18 & 96.02 & 84.19 & 68.13 & 75.13 \\
AgentSquare    & 87.62 & 48.51 & 97.77 & 89.08 & 78.46 & 80.29 \\
AFlow          & 91.16 & 51.28 & 96.22 & 90.93 & 81.67 & 82.25 \\
MaAS           & \underline{92.30} & \underline{51.82} & \textbf{98.80} & \underline{92.85} & \underline{82.17} & \underline{83.59} \\
\midrule
\graft{} & \textbf{95.04} & \textbf{62.89} & \underline{97.9} & \textbf{94.66} & \textbf{86.70} & \textbf{87.44} \\
\bottomrule
\end{tabular}%
}
\caption{MaAS setup (gpt-4o-mini executor)}
\label{tab:main}
\end{subtable}

\par\vspace{6pt}

\begin{subtable}{\columnwidth}
\centering
\setlength{\tabcolsep}{4pt}\small
\resizebox{\columnwidth}{!}{%
\begin{tabular}{lccccccc}
\toprule
Method & GSM8K & MATH & HotpotQA & DROP & MMLU-Pro & GPQA & Avg. \\
\midrule
IO         & 90.0 & 40.2 & 57.4 & 75.1 & 57.8 & 44.8 & 60.9 \\
CoT        & 95.2 & 43.2 & 45.1 & 81.5 & 77.4 & 65.6 & 68.0 \\
CoT-SC     & 96.0 & 45.0 & 21.8 & 49.9 & 24.3 & 65.8 & 50.5 \\
ADAS       & 96.0 & 44.1 & 73.3 & 81.2 & 80.1 & 64.0 & 73.1 \\
MaAS       & 96.4 & 41.3 & 76.3 & 84.2 & 82.0 & 55.2 & 72.6 \\
AFlow      & \underline{96.5} & 60.1 & 63.7 & 89.2 & \underline{82.3} & 65.3 & 76.2 \\
BayesFlow  & 96.0 & \underline{69.4} & \underline{77.5} & \underline{90.8} & 81.8 & \underline{69.2} & \underline{80.8} \\
\midrule
\graft{}  & \textbf{97.2} & \textbf{76.8} & \textbf{80.0} & \textbf{92.7} & \textbf{83.4} & \textbf{74.2} & \textbf{84.1} \\
\bottomrule
\end{tabular}%
}
\caption{BayesFlow setup (Claude-Sonnet executor)}
\label{tab:bayesflow}
\end{subtable}

\caption{
Performance across two evaluation setups: (a) MaAS and (b) BayesFlow.
Bold = best, \underline{underline} = runner-up.
GPQA absolute values should be read with caution due to possible contamination. MultiArith is measured with the GSM8K-base workflow.}
\label{tab:results}
\end{table}

%% file: tabs/tab_ablcomp.tex
\begin{table}[t]
\centering
\setlength{\tabcolsep}{4pt}
\resizebox{\columnwidth}{!}{%
\begin{tabular}{lcccccc}
\toprule
Configuration & GSM8K & MATH & MultiA. & HumanE. & MBPP & Avg. \\
\midrule
Full (\graft{})       & 95.04 & 62.89 & 97.90 & 94.66 & 86.70 & \textbf{87.44} \\
$-$ local grafting    & 93.27 & 56.86 & 98.11 & 86.26 & 82.21 & 83.34 \\
$-$ label-free proxy  & 93.08 & 57.13 & 97.11 & 89.57 & 82.89 & 83.96 \\
$-$ coupling guard    & 94.95 & 61.73 & 98.22 & 91.60 & 87.19 & 86.74 \\
\midrule
global $P{=}1$        & 94.85 & 62.83 & 97.89 & 93.64 & 87.29 & 87.30 \\
global $P{=}3$        & 95.11 & 62.34 & 98.22 & 90.33 & 86.61 & 86.52 \\
per-task $P^{\ast}$   & 95.04 & 62.89 & 97.90 & 94.66 & 86.70 & \textbf{87.44} \\
\bottomrule
\end{tabular}}
\caption{Component ablation (top) and $P$-strategy comparison (bottom) on the MaAS suite
(five benchmarks, three-run averages). Bold = best Avg.}
\label{tab:abl-comp}
\end{table}

%% file: body/analysis.tex
\section{Analysis}

\subsection{Fidelity of the Label-Free Proxy}
\input{tabs/tab_qa}
Table~\ref{tab:qa} analyzes the runtime statistics of \graft{} across benchmarks (RQ3).
The Spearman correlation between the label-free proxy $Q$ and answer correctness ranges from 0.23 to 0.62 for code and math tasks, whereas it is at most 0.17 for all QA and knowledge-intensive tasks.
This indicates that the proxy is less effective at distinguishing good candidates in QA and knowledge-intensive tasks, which rely on groundedness rather than directly verifiable execution outcomes.

\subsection{Effect of Memory Amortization}
\input{rsc/amortize}

\input{rsc/exe_transfer}
\input{tabs/tab_case_qa_graft}

The memory reuse rate (Reuse) in Table~\ref{tab:qa} confirms that winning configurations are reused in practice (RQ4). 
For GSM8K and MATH, which have relatively uniform input types, reuse rates range from 0.86 to 0.92, suggesting that local search is rarely needed.
Reuse rates decrease for code, QA, and knowledge-intensive tasks with more diverse input types.
The concurrent reductions in token usage and the number of passes as reuse rates increase indicate that local search costs are amortized across inputs, allowing the search process to converge for uniform input types.

Figure~\ref{fig:amortize} shows that search cost decreases by approximately 50\% after the initial decile as configuration memory accumulates reusable workflows, while configuration reuse increases and stabilizes. 
Reuse reaches 1.2, 0.8, and 0.5 regions/query for MBPP, DROP, and GSM8K, respectively, demonstrating that amortization generalizes across task types.

\subsection{Executor Transfer}

Figure~\ref{fig:exec} compares the same $W_{c^{*}}$ searched with gpt-4o-mini across different executors without re-optimizing the workflow.
As the executor is replaced with increasingly capable models, from gpt-4o-mini to Haiku and then to Sonnet, accuracy generally increases monotonically, with particularly large gains on challenging reasoning tasks such as GPQA, MMLU-Pro, and MATH.
In contrast, the gains beyond Haiku are small for GSM8K and MultiArith, where performance is already saturated, and for HotpotQA, where retrieval remains the bottleneck and token-level F1 is also saturated.
This cost-accuracy operating point is achieved by changing only the executor backbone without re-optimizing the workflow, showing that the optimized workflow functions as an adaptive execution policy.

\subsection{Operator Trace of a Multi-hop QA Correction}
\label{app:case-qa-graft}

Table~\ref{tab:case-qa-graft} shows the results of inference-time grafting applied to the globally optimized workflow.
The correct answer to the question ``The Distribution of Industry Act was passed by a man who was prime minister when?'' is Prime Minister Clement Attlee's term in office, \emph{1945 to 1951}.
The plan (R1) and evidence (R2) regions in both pipelines retrieve the same evidence, ``Attlee\ldots served as PM from 1945 to 1951'', and the distractor ``Distribution of Industry Act \emph{1950}.''
The single \textsf{CoT} chain in Frozen mistakes the Act's year, $1950$, for the prime minister's term and concludes that ``the year of their term is 1950.''
This error then propagates through the format region to the final answer, $1950$.
In contrast, \graft{} optimizes the reason region to \textsf{SC5+SCcheck+Revise} based on its low label-free $Q$: Extract five independent chains to form a self-consistency majority, then refine the result using self-check and revision.
All five chains agree on ``Prime Minister from 1945 to 1951'', reaching $Q{=}1.00$, and the final answer is corrected to \emph{1945 to 1951}.

%% file: tabs/tab_qa.tex
\begin{table}[t]
\centering
\setlength{\tabcolsep}{3.5pt}
\small
\begin{tabular}{llrrrr}
\toprule
Dataset & Metric & Score & Tokens & Reuse & $\rho$ \\
\midrule
GSM8K       & Solve rate & 95.04 & \phantom{0}3{,}506  & 0.86 & 0.35 \\
MATH        & Solve rate & 62.89 & \phantom{0}9{,}039  & 0.92 & 0.57 \\
MultiArith  & solve rate & 97.90 & \phantom{0}1{,}582  & 0.97 & 0.23 \\
HumanEval   & Pass@1     & 94.66 & \phantom{0}3{,}455  & 0.42 & 0.31 \\
MBPP        & Pass@1     & 86.70 & \phantom{0}1{,}955  & 0.63 & 0.62 \\
HotpotQA    & F1         & 74.58 & 39{,}947 & 0.42 & 0.06 \\
DROP        & F1         & 85.55 & \phantom{0}9{,}082 & 0.24 & 0.12 \\
MMLU-Pro    & Accuracy   & 62.87 & 12{,}076 & 0.20 & 0.17 \\
GPQA        & Accuracy   & 37.74 & 20{,}312 & 0.17 & 0.04 \\
\bottomrule
\end{tabular}
\caption{\graft{} operating statistics across all benchmarks. $\rho$ is the Spearman correlation between the label-free proxy $Q$ and answer correctness; Reuse is the configuration-memory hit rate. MultiArith is measured with the GSM8K-based workflow.
}
\label{tab:qa}
\end{table}

%% file: rsc/amortize.tex
\begin{figure}[t]
\centering
\includegraphics[width=\columnwidth]{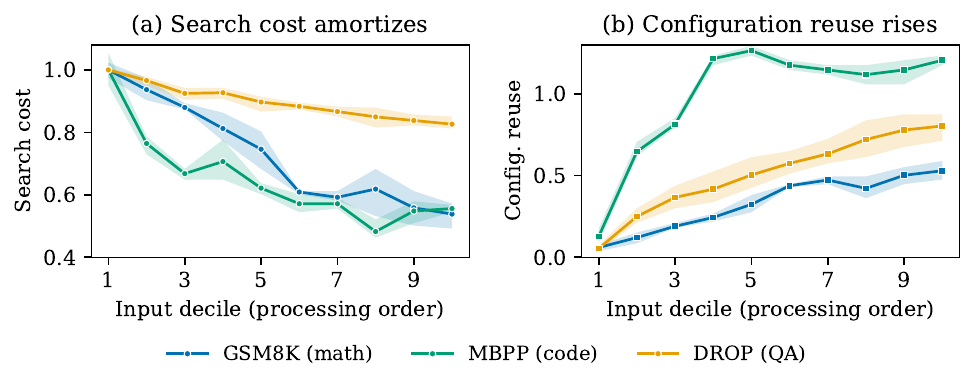}
\caption{
Per-query search cost (left, relative to the first decile) and configuration-memory reuse (right, the number of regions reused per query) over the processing order (decile). Lines are the mean over three runs and shading is the min–max range.
}
\label{fig:amortize}
\end{figure}

%% file: rsc/exe_transfer.tex
\begin{figure}[t]
\centering
\includegraphics[width=0.9\columnwidth]{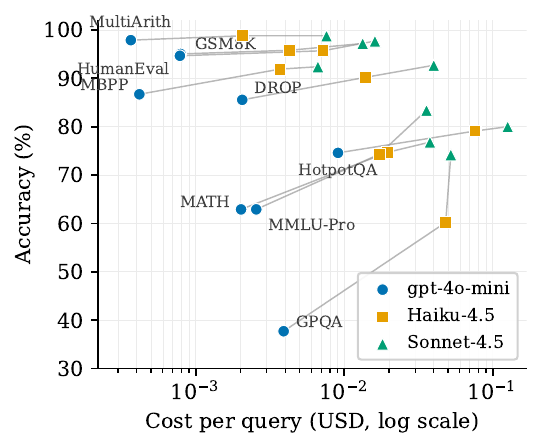}
\caption{Per-dataset cost-accuracy when the executor is swapped gpt-4o-mini $\to$ Haiku $\to$ Sonnet on the same $W_{c^{*}}$. Each line connects the three executor points of one
dataset.}
\label{fig:exec}
\end{figure}

%% file: tabs/tab_case_qa_graft.tex
\begin{table*}[t]
\centering
\setlength{\tabcolsep}{5pt}\small
\begin{tabular}{p{0.115\textwidth} p{0.395\textwidth} p{0.395\textwidth}}
\toprule
\textbf{Query} & \multicolumn{2}{p{0.83\textwidth}}{\emph{The Distribution of Industry Act was passed by a man who was prime minister when?}} \\
\textbf{Ground truth} & \multicolumn{2}{p{0.83\textwidth}}{\emph{1945 to 1951} \;(PM Clement Attlee's term)} \\
\midrule
Region (role) & \textbf{Frozen $W_{c^{*}}$} (operator: output) $\Rightarrow$ \textcolor{red}{wrong} &
\graft{} (operator: output) $\Rightarrow$ \textcolor{green!55!black}{correct} \\
\midrule
R1 plan &
\textsf{Planner}: ``\emph{2.\ Research the year the act was passed.\ 3.\ Determine who was PM
during that year.\ 4.\ Provide the PM's name and the year of their term.}'' &
\textsf{Planner} (same decomposition). \\
\addlinespace
R2 evidence &
\textsf{Retrieve+Read}: ``\emph{The Distribution of Industry Act 1950 was passed by the Labour
government of Clement Attlee.\ Attlee served as Prime Minister \ldots from 1945 to 1951.}'' &
\textsf{MHop2+Read}: same evidence retrieved ($Q{=}0.96$). \\
\addlinespace
R3 reason &
\textsf{CoT} (single chain): ``\emph{\ldots the year the act was passed is 1950.\ \ldots the name
of the prime minister is Clement Attlee and the year of their term is \textbf{1950}.}''
$\Rightarrow$ \textcolor{red}{Answer: 1950} &
\underline{\textsf{SC5+SCcheck+Revise}} \textsc{[grafted]} (5 chains): ``\emph{\ldots the act was
passed by Clement Attlee, who was Prime Minister \textbf{from 1945 to 1951}.}'' all 5 agree,
$Q{=}1.00$. \\
\midrule
Final answer & \textcolor{red}{\textbf{1950}} \;(confuses the act's year with the PM's term) &
\textcolor{green!55!black}{\textbf{1945 to 1951}} \\
\bottomrule
\end{tabular}
\caption{A multi-hop QA case where inference-time local optimization corrects the answer, with verbatim input/output excerpts for each region operator.}
\label{tab:case-qa-graft}
\end{table*}

%% file: body/conclusion.tex
\section{Conclusion}

We present \graft{}, a training-free framework that grafts region-level local search onto a task-specific global workflow for per-input adaptation at inference time.
\graft{} evaluates candidates without ground-truth labels using self-consistency, groundedness, and verifier signals, accepts replacements only when a coupling guard preserves cross-boundary consistency, and caches winning configurations for reuse.
Across five benchmarks with GPT-4o-mini, \graft{} achieves an average score of 87.44, outperforming the supernet-based MaAS by 3.85 points and the search-based AFlow by 5.19 points.
Ablation results show that local grafting yields consistent gains across global workflow sources and reduces search cost on tasks with high memory reuse.

%% file: aaai2027.bib
@misc{zhao2026surveylargelanguagemodels,
      title={A Survey of Large Language Models}, 
      author={Wayne Xin Zhao and Kun Zhou and Junyi Li and Tianyi Tang and Xiaolei Wang and Yupeng Hou and Yingqian Min and Beichen Zhang and Junjie Zhang and Zican Dong and Yifan Du and Chen Yang and Yushuo Chen and Zhipeng Chen and Jinhao Jiang and Ruiyang Ren and Yifan Li and Xinyu Tang and Zikang Liu and Peiyu Liu and Jian-Yun Nie and Ji-Rong Wen},
      year={2026},
      eprint={2303.18223},
      archivePrefix={arXiv},
      primaryClass={cs.CL},
      url={https://arxiv.org/abs/2303.18223}, 
}

@misc{minaee2025largelanguagemodelssurvey,
      title={Large Language Models: A Survey}, 
      author={Shervin Minaee and Tomas Mikolov and Narjes Nikzad and Meysam Chenaghlu and Richard Socher and Xavier Amatriain and Jianfeng Gao},
      year={2025},
      eprint={2402.06196},
      archivePrefix={arXiv},
      primaryClass={cs.CL},
      url={https://arxiv.org/abs/2402.06196}, 
}

@ARTICLE{6145622,
  author={Browne, Cameron B. and Powley, Edward and Whitehouse, Daniel and Lucas, Simon M. and Cowling, Peter I. and Rohlfshagen, Philipp and Tavener, Stephen and Perez, Diego and Samothrakis, Spyridon and Colton, Simon},
  journal={IEEE Transactions on Computational Intelligence and AI in Games}, 
  title={A Survey of Monte Carlo Tree Search Methods}, 
  year={2012},
  volume={4},
  number={1},
  pages={1-43},
  doi={10.1109/TCIAIG.2012.2186810}}

@article{Wang_2024,
   title={A survey on large language model based autonomous agents},
   volume={18},
   ISSN={2095-2236},
   url={http://dx.doi.org/10.1007/s11704-024-40231-1},
   DOI={10.1007/s11704-024-40231-1},
   number={6},
   journal={Frontiers of Computer Science},
   publisher={Springer Science and Business Media LLC},
   author={Wang, Lei and Ma, Chen and Feng, Xueyang and Zhang, Zeyu and Yang, Hao and Zhang, Jingsen and Chen, Zhiyuan and Tang, Jiakai and Chen, Xu and Lin, Yankai and Zhao, Wayne Xin and Wei, Zhewei and Wen, Jirong},
   year={2024},
   month=Mar }

@misc{xi2023risepotentiallargelanguage,
      title={The Rise and Potential of Large Language Model Based Agents: A Survey}, 
      author={Zhiheng Xi and Wenxiang Chen and Xin Guo and Wei He and Yiwen Ding and Boyang Hong and Ming Zhang and Junzhe Wang and Senjie Jin and Enyu Zhou and Rui Zheng and Xiaoran Fan and Xiao Wang and Limao Xiong and Yuhao Zhou and Weiran Wang and Changhao Jiang and Yicheng Zou and Xiangyang Liu and Zhangyue Yin and Shihan Dou and Rongxiang Weng and Wensen Cheng and Qi Zhang and Wenjuan Qin and Yongyan Zheng and Xipeng Qiu and Xuanjing Huang and Tao Gui},
      year={2023},
      eprint={2309.07864},
      archivePrefix={arXiv},
      primaryClass={cs.AI},
      url={https://arxiv.org/abs/2309.07864}, 
}

@misc{liu2023llmpempoweringlargelanguage,
      title={LLM+P: Empowering Large Language Models with Optimal Planning Proficiency}, 
      author={Bo Liu and Yuqian Jiang and Xiaohan Zhang and Qiang Liu and Shiqi Zhang and Joydeep Biswas and Peter Stone},
      year={2023},
      eprint={2304.11477},
      archivePrefix={arXiv},
      primaryClass={cs.AI},
}

@article{Hatalis_Christou_Myers_Jones_Lambert_Amos-Binks_Dannenhauer_Dannenhauer_2024, title={Memory Matters: The Need to Improve Long-Term Memory in LLM-Agents}, volume={2}, url={https://ojs.aaai.org/index.php/AAAI-SS/article/view/27688}, DOI={10.1609/aaaiss.v2i1.27688}, abstractNote={In this paper, we provide a review of the current efforts to develop LLM agents, which are autonomous agents that leverage large language models. We examine the memory management approaches used in these agents. One crucial aspect of these agents is their long-term memory, which is often implemented using vector databases. We describe how vector databases are utilized to store and retrieve information in LLM agents. Moreover we highlight open problems, such as the separation of different types of memories and the management of memory over the agent’s lifetime. Lastly, we propose several topics for future research to address these challenges and further enhance the capabilities of LLM agents, including the use of metadata in procedural and semantic memory and the integration of external knowledge sources with vector databases.}, number={1}, journal={Proceedings of the AAAI Symposium Series}, author={Hatalis, Kostas and Christou, Despina and Myers, Joshua and Jones, Steven and Lambert, Keith and Amos-Binks, Adam and Dannenhauer, Zohreh and Dannenhauer, Dustin}, year={2024}, month={Jan.}, pages={277–280} }

@misc{aflow,
      title={AFlow: Automating Agentic Workflow Generation}, 
      author={Jiayi Zhang and Jinyu Xiang and Zhaoyang Yu and Fengwei Teng and Xionghui Chen and Jiaqi Chen and Mingchen Zhuge and Xin Cheng and Sirui Hong and Jinlin Wang and Bingnan Zheng and Bang Liu and Yuyu Luo and Chenglin Wu},
      year={2025},
      eprint={2410.10762},
      archivePrefix={arXiv},
      primaryClass={cs.AI},
}

@misc{A2Flow,
      title={$A^2Flow:$ Automating Agentic Workflow Generation via Self-Adaptive Abstraction Operators}, 
      author={Mingming Zhao and Xiaokang Wei and Yuanqi Shao and Kaiwen Zhou and Lin Yang and Siwei Rao and Junhui Zhan and Zhitang Chen},
      year={2025},
      eprint={2511.20693},
      archivePrefix={arXiv},
      primaryClass={cs.AI},
}

@misc{AgentSquare,
      title={AgentSquare: Automatic LLM Agent Search in Modular Design Space}, 
      author={Yu Shang and Yu Li and Keyu Zhao and Likai Ma and Jiahe Liu and Fengli Xu and Yong Li},
      year={2025},
      eprint={2410.06153},
      archivePrefix={arXiv},
      primaryClass={cs.CL},
}

@misc{AgentSwift,
      title={AgentSwift: Efficient LLM Agent Design via Value-guided Hierarchical Search}, 
      author={Yu Li and Lehui Li and Zhihao Wu and Qingmin Liao and Jianye Hao and Kun Shao and Fengli Xu and Yong Li},
      year={2025},
      eprint={2506.06017},
      archivePrefix={arXiv},
      primaryClass={cs.CL},
}

@inproceedings{FusionFlow,
    title = "{F}usion{F}low: Enabling Deep Structural Exploration for Automated Agentic Workflow Generation",
    author = "Wang, Xiang  and
      Yang, Zongtao  and
      Hong, Zhuojian  and
      Zhang, Shuhao  and
      Wei, Wei",
    editor = "Liakata, Maria  and
      Moreira, Viviane P.  and
      Zhang, Jiajun  and
      Jurgens, David",
    booktitle = "Proceedings of the 64th Annual Meeting of the {A}ssociation for {C}omputational {L}inguistics (Volume 1: Long Papers)",
    month = jul,
    year = "2026",
    address = "San Diego, California, United States",
    publisher = "Association for Computational Linguistics",
    pages = "27718--27760",
    ISBN = "979-8-89176-390-6",
}

@misc{EvoFlow,
      title={EvoFlow: Evolving Diverse Agentic Workflows On The Fly}, 
      author={Guibin Zhang and others},
      year={2025},
      eprint={2502.07373},
      archivePrefix={arXiv},
      primaryClass={cs.LG},
}

@misc{ScoreFlow,
      title={ScoreFlow: Mastering LLM Agent Workflows via Score-based Preference Optimization}, 
      author={Yinjie Wang and Ling Yang and Guohao Li and Mengdi Wang and Bryon Aragam},
      year={2025},
      eprint={2502.04306},
      archivePrefix={arXiv},
      primaryClass={cs.CL},
}

@misc{DyLAN,
      title={A Dynamic LLM-Powered Agent Network for Task-Oriented Agent Collaboration}, 
      author={Zijun Liu and Yanzhe Zhang and Peng Li and Yang Liu and Diyi Yang},
      year={2024},
      eprint={2310.02170},
      archivePrefix={arXiv},
      primaryClass={cs.CL},
}

@misc{DyFlow,
      title={DyFlow: Dynamic Workflow Framework for Agentic Reasoning}, 
      author={Yanbo Wang and Zixiang Xu and Yue Huang and Xiangqi Wang and Zirui Song and Lang Gao and Chenxi Wang and Xiangru Tang and Yue Zhao and Arman Cohan and Xiangliang Zhang and Xiuying Chen},
      year={2025},
      eprint={2509.26062},
      archivePrefix={arXiv},
      primaryClass={cs.CL},
      url={https://arxiv.org/abs/2509.26062}, 
}

@misc{Flow,
      title={Flow: Modularized Agentic Workflow Automation}, 
      author={Boye Niu and Yiliao Song and Kai Lian and Yifan Shen and Yu Yao and Kun Zhang and Tongliang Liu},
      year={2025},
      eprint={2501.07834},
      archivePrefix={arXiv},
      primaryClass={cs.AI},
}

@inproceedings{HFlow,
    title = "Evolving Agentic Workflow Driven by Human-Agent Collaboration",
    author = "Liu, Yuxin  and
      others",
    editor = "Liakata, Maria  and
      Moreira, Viviane P.  and
      Zhang, Jiajun  and
      Jurgens, David",
    booktitle = "Findings of the {A}ssociation for {C}omputational {L}inguistics: {ACL} 2026",
    month = jul,
    year = "2026",
    address = "San Diego, California, United States",
    publisher = "Association for Computational Linguistics",
    pages = "24960--24969",
    ISBN = "979-8-89176-395-1",
}

@misc{adas,
      title={Automated Design of Agentic Systems}, 
      author={Shengran Hu and Cong Lu and Jeff Clune},
      year={2025},
      eprint={2408.08435},
      archivePrefix={arXiv},
      primaryClass={cs.AI},
      url={https://arxiv.org/abs/2408.08435}, 
}

@inproceedings{sese,
  author       = {Richard Johnson and
                  David Pearson and
                  Keshav Pingali},
  editor       = {Vivek Sarkar and
                  Barbara G. Ryder and
                  Mary Lou Soffa},
  title        = {The Program Structure Tree: Computing Control Regions in Linear Time},
  booktitle    = {Proceedings of the {ACM} SIGPLAN'94 Conference on Programming Language
                  Design and Implementation (PLDI), Orlando, Florida, USA, June 20-24,
                  1994},
  pages        = {171--185},
  publisher    = {{ACM}},
  year         = {1994},
  url          = {https://doi.org/10.1145/178243.178258},
  doi          = {10.1145/178243.178258}
}

@article{mast,
  author  = {Mert Cemri and Melissa Z. Pan and Shuyi Yang and others},
  title   = {Why Do Multi-Agent {LLM} Systems Fail?},
  journal = {arXiv preprint arXiv:2503.13657},
  year    = {2025}
}

@inproceedings{maas,
  author    = {Guibin Zhang and Luyang Niu and Junfeng Fang and Kun Wang and Lei Bai and Xiang Wang},
  title     = {Multi-agent Architecture Search via Agentic Supernet},
  booktitle = {International Conference on Machine Learning (ICML)},
  year      = {2025}
}

@inproceedings{math_ds,
  author    = {Dan Hendrycks and Collin Burns and Saurav Kadavath and Akul Arora and Steven Basart and Eric Tang and Dawn Song and Jacob Steinhardt},
  title     = {Measuring Mathematical Problem Solving With the {MATH} Dataset},
  booktitle = {NeurIPS Datasets and Benchmarks},
  year      = {2021}
}

@article{mbpp,
  author  = {Jacob Austin and Augustus Odena and Maxwell Nye and others},
  title   = {Program Synthesis with Large Language Models},
  journal = {arXiv preprint arXiv:2108.07732},
  year    = {2021}
}

@inproceedings{multiarith,
  author    = {Subhro Roy and Dan Roth},
  title     = {Solving General Arithmetic Word Problems},
  booktitle = {Empirical Methods in Natural Language Processing (EMNLP)},
  year      = {2015}
}

@inproceedings{selfconsistency,
  author    = {Xuezhi Wang and Jason Wei and Dale Schuurmans and others},
  title     = {Self-Consistency Improves Chain of Thought Reasoning in Language Models},
  booktitle = {International Conference on Learning Representations (ICLR)},
  year      = {2023}
}

@inproceedings{debate_icml24,
author = {Du, Yilun and Li, Shuang and Torralba, Antonio and Tenenbaum, Joshua B. and Mordatch, Igor},
title = {Improving factuality and reasoning in language models through multiagent debate},
year = {2024},
publisher = {JMLR.org},
booktitle = {Proceedings of the 41st International Conference on Machine Learning},
articleno = {467},
numpages = {31},
location = {Vienna, Austria},
series = {ICML'24}
}

@inproceedings{reflexion_neurips23,
  author    = {Noah Shinn and Federico Cassano and Ashwin Gopinath and Karthik Narasimhan and Shunyu Yao},
  title     = {Reflexion: Language Agents with Verbal Reinforcement Learning},
  booktitle = {Advances in Neural Information Processing Systems (NeurIPS)},
  year      = {2023}
}

@inproceedings{selfrefine,
  author    = {Aman Madaan and Niket Tandon and Prakhar Gupta and others},
  title     = {Self-Refine: Iterative Refinement with Self-Feedback},
  booktitle = {Advances in Neural Information Processing Systems (NeurIPS)},
  year      = {2023}
}

@inproceedings{gptswarm,
  author    = {Mingchen Zhuge and Wenyi Wang and Louis Kirsch and Francesco Faccio and Dmitrii Khizbullin and J{\"u}rgen Schmidhuber},
  title     = {{GPTSwarm}: Language Agents as Optimizable Graphs},
  booktitle = {International Conference on Machine Learning (ICML)},
  year      = {2024}
}

@inproceedings{rag,
  author    = {Patrick Lewis and Ethan Perez and Aleksandra Piktus and others},
  title     = {Retrieval-Augmented Generation for Knowledge-Intensive {NLP} Tasks},
  booktitle = {Advances in Neural Information Processing Systems (NeurIPS)},
  year      = {2020}
}

@article{autogen,
  author  = {Qingyun Wu and Gagan Bansal and Jieyu Zhang and others},
  title   = {{AutoGen}: Enabling Next-Gen {LLM} Applications via Multi-Agent Conversation},
  journal = {arXiv preprint arXiv:2308.08155},
  year    = {2023}
}

@misc{camel,
      title={CAMEL: Communicative Agents for "Mind" Exploration of Large Language Model Society}, 
      author={Guohao Li and Hasan Abed Al Kader Hammoud and Hani Itani and Dmitrii Khizbullin and Bernard Ghanem},
      year={2023},
      eprint={2303.17760},
      archivePrefix={arXiv},
      primaryClass={cs.AI},
}

@misc{metagpt,
      title={MetaGPT: Meta Programming for A Multi-Agent Collaborative Framework}, 
      author={Sirui Hong and Mingchen Zhuge and Jiaqi Chen and Xiawu Zheng and Yuheng Cheng and Ceyao Zhang and Jinlin Wang and Zili Wang and Steven Ka Shing Yau and Zijuan Lin and Liyang Zhou and Chenyu Ran and Lingfeng Xiao and Chenglin Wu and Jürgen Schmidhuber},
      year={2024},
      eprint={2308.00352},
      archivePrefix={arXiv},
      primaryClass={cs.AI},
      url={https://arxiv.org/abs/2308.00352}, 
}

@misc{agentverse,
      title={AgentVerse: Facilitating Multi-Agent Collaboration and Exploring Emergent Behaviors}, 
      author={Weize Chen and Yusheng Su and Jingwei Zuo and Cheng Yang and Chenfei Yuan and Chi-Min Chan and Heyang Yu and Yaxi Lu and Yi-Hsin Hung and Chen Qian and Yujia Qin and Xin Cong and Ruobing Xie and Zhiyuan Liu and Maosong Sun and Jie Zhou},
      year={2023},
      eprint={2308.10848},
      archivePrefix={arXiv},
      primaryClass={cs.CL},
}

@article{humaneval,
  author  = {Mark Chen and Jerry Tworek and Heewoo Jun and others},
  title   = {Evaluating Large Language Models Trained on Code},
  journal = {arXiv preprint arXiv:2107.03374},
  year    = {2021}
}

@inproceedings{cot_neurips22,
  author    = {Jason Wei and Xuezhi Wang and Dale Schuurmans and others},
  title     = {Chain-of-Thought Prompting Elicits Reasoning in Large Language Models},
  booktitle = {Advances in Neural Information Processing Systems (NeurIPS)},
  year      = {2022}
}

@inproceedings{hotpotqa,
  author    = {Zhilin Yang and Peng Qi and Saizheng Zhang and others},
  title     = {{HotpotQA}: A Dataset for Diverse, Explainable Multi-hop Question Answering},
  booktitle = {Empirical Methods in Natural Language Processing (EMNLP)},
  year      = {2018}
}

@inproceedings{drop,
  author    = {Dheeru Dua and Yizhong Wang and Pradeep Dasigi and others},
  title     = {{DROP}: A Reading Comprehension Benchmark Requiring Discrete Reasoning Over Paragraphs},
  booktitle = {North American Chapter of the Association for Computational Linguistics (NAACL)},
  year      = {2019}
}

@article{gsm8k,
  author  = {Karl Cobbe and Vineet Kosaraju and Mohammad Bavarian and others},
  title   = {Training Verifiers to Solve Math Word Problems},
  journal = {arXiv preprint arXiv:2110.14168},
  year    = {2021}
}

@article{ucb,
  author       = {Peter Auer and Nicol{\`o} Cesa-Bianchi and Paul Fischer},
  title        = {Finite-time Analysis of the Multiarmed Bandit Problem},
  journal      = {Machine Learning},
  volume       = {47},
  number       = {2--3},
  pages        = {235--256},
  year         = {2002}
}

@inproceedings{mmlupro,
  author    = {Yubo Wang and Xueguang Ma and Ge Zhang and others},
  title     = {{MMLU-Pro}: A More Robust and Challenging Multi-Task Language Understanding Benchmark},
  booktitle = {Advances in Neural Information Processing Systems (NeurIPS), Datasets and Benchmarks Track},
  year      = {2024}
}

@inproceedings{gpqa,
  author    = {David Rein and Betty Li Hou and Asa Cooper Stickland and others},
  title     = {{GPQA}: A Graduate-Level Google-Proof Q\&A Benchmark},
  booktitle = {Conference on Language Modeling (COLM)},
  year      = {2024}
}

@inproceedings{yuan-etal-2026-bayesflow,
    title = "{B}ayes{F}low: A Probability Inference Framework for Meta-Agent Assisted Workflow Generation",
    author = "Yuan, Bo  and
      Zhou, Yun  and
      Xu, Zhichao  and
      Ramnath, Kiran  and
      Feng, Aosong  and
      Srinivasan, Balasubramaniam",
    editor = "Demberg, Vera  and
      Inui, Kentaro  and
      Marquez, Llu{\'i}s",
    booktitle = "Findings of the {A}ssociation for {C}omputational {L}inguistics: {EACL} 2026",
    month = mar,
    year = "2026",
    address = "Rabat, Morocco",
    publisher = "Association for Computational Linguistics",
    url = "https://aclanthology.org/2026.findings-eacl.165/",
    doi = "10.18653/v1/2026.findings-eacl.165",
    pages = "3151--3179",
    ISBN = "979-8-89176-386-9"
}
